\documentclass[letterpaper, 10 pt, journal, twoside]{IEEEtran} 

\IEEEoverridecommandlockouts                              % This command is only needed if 
\usepackage{todonotes}
\usepackage{csquotes}
\usepackage{amsfonts}
\usepackage{amssymb}
\usepackage{amsfonts}
\usepackage{amsmath}
\usepackage{caption}
\usepackage{subcaption}
\usepackage{graphicx}
\usepackage{svg}
\usepackage{cite}
\usepackage[hang,flushmargin,symbol]{footmisc}
\usepackage{bm}
\usepackage{flushend}

\usepackage{booktabs}
\usepackage{xspace}
\usepackage{array}
\usepackage{tabularx}
\usepackage{enumitem}
\usepackage{siunitx}
\usepackage{gensymb}
\usepackage{microtype}

\usepackage{xcolor}
\definecolor{red}{RGB}{255, 0, 0}   % red
\definecolor{orange}{RGB}{255, 77, 0}   % Orange
\definecolor{green}{RGB}{0, 128, 0}   % Green
\definecolor{purple}{RGB}{160, 32, 240}   % Purple
\definecolor{lightblue}{RGB}{52, 155, 235}   % light blue
\definecolor{darkmagenta}{RGB}{204, 51, 139}
\newcommand{\rebuttal}[0]{}

\renewcommand{\baselinestretch}{0.99}
\usepackage{hyperref}
\hypersetup{breaklinks,colorlinks}

\usepackage[capitalize]{cleveref}
\crefname{figure}{Fig.}{Figs.}
\Crefname{figure}{Figure}{Figures}
\crefname{section}{Sec.}{Secs.}
\Crefname{section}{Section}{Sections}
\Crefname{table}{Table}{Tables}
\crefname{table}{Tab.}{Tabs.}
\crefname{algorithm}{Algo.}{Algos.}
\Crefname{algorithm}{Algorithm}{Algorithms}
\crefname{appendix}{Sec.}{Secs.}
\Crefname{appendix}{Section}{Sections}

\newcommand{\ourName}{\mbox{CenterGrasp}} % \mbox prevents breaking
\newcommand{\ourNameFull}{Object-Aware Implicit Representation Learning for Simultaneous Shape Reconstruction and 6-DoF Grasp Estimation}

\newcommand{\latentVariable}{\boldsymbol{z}}
\newcommand{\allLatentVariables}{\boldsymbol{Z}}
\newcommand{\latentVariablesDim}{D_l}

\newcommand{\realNumber}{\mathbb{R}}

\newcommand{\sdfValue}{s}
\newcommand{\allSdfValues}{\boldsymbol{S}}
\newcommand{\spaceCoordinate}{\boldsymbol{x}}

\newcommand{\graspOutput}{\tilde{\boldsymbol{g}}}
\newcommand{\graspDelta}{\delta\boldsymbol{t}}
\newcommand{\graspRotation}{\boldsymbol{r}}
\newcommand{\graspRotationFull}{\boldsymbol{R}}
\newcommand{\transformation}{\boldsymbol{T}}
\newcommand{\grasp}{\boldsymbol{g}}
\newcommand{\allGrasps}{\boldsymbol{G}}

\newcommand{\lossGeneralFunction}{\mathcal{L}}
\newcommand{\lossScaling}{\delta}

\newcommand{\clampThreshold}{c}
\newcommand{\gripperPoints}{\boldsymbol{v}} % After rebuttal
\newcommand{\gripperPointsFlipped}{\tilde{\gripperPoints}}

\newcommand{\epoch}{e}

\title{%\LARGE \bf
\ourName: \ourNameFull
}

\author{Eugenio Chisari$^{1}$, Nick Heppert$^{1}$, Tim Welschehold$^{1}$, Wolfram Burgard$^{2}$, Abhinav Valada$^{1}$\vspace{-0.4cm}% <-this % stops a space
\thanks{Manuscript received: December, 13, 2023; Revised February, 29, 2024; Accepted April, 2, 2024.}%Use only for final RAL version
\thanks{This paper was recommended for publication by Editor M. Vincze upon evaluation of the Associate Editor and Reviewers' comments.
This work was funded by the Carl Zeiss Foundation with the ReScaLe project.} %Use only for final RAL version
\thanks{$^{1}$ Department of Computer Science, University of Freiburg, Germany}% <-this % stops a space
\thanks{$^{2}$ Department of Engineering, University of Technology Nuremberg, Germany}% <-this % stops a space
\thanks{Digital Object Identifier (DOI): see top of this page.}
}

\begin{document}
\maketitle
% \thispagestyle{empty}
% \pagestyle{empty}
% Paper headers
\markboth{IEEE Robotics and Automation Letters. Preprint Version. Accepted April, 2024}
{Chisari \MakeLowercase{\textit{et al.}}: CenterGrasp} 
% Use only for final RAL version
%===============================================================================
\begin{abstract}
Reliable object grasping is a crucial capability for autonomous robots. However, many existing grasping approaches focus on general clutter removal without explicitly modeling objects and thus only relying on the visible local geometry. We introduce \ourName{}, a novel framework that combines object awareness and holistic grasping. \ourName{} learns a general object prior by encoding shapes and valid grasps in a continuous latent space. It consists of an \mbox{RGB-D} image encoder that leverages recent advances to detect objects and infer their pose and latent code, and a decoder to predict shape and grasps for each object in the scene. We perform extensive experiments on simulated as well as real-world cluttered scenes and demonstrate strong scene reconstruction and 6-DoF grasp-pose estimation performance. Compared to the state of the art, \ourName{} achieves an improvement of 38.5 mm in shape reconstruction and 33 percentage points on average in grasp success.
We make the code and trained models publicly available at \url{http://centergrasp.cs.uni-freiburg.de}.
\end{abstract}
% Keywords appear just beneath the abstract. Use only for final RAL version. 
%===============================================================================

\section{Introduction}

% Describe the problem: 6-DoF grasp-pose estimation
\IEEEPARstart{G}{rasp-pose} estimation is a high-dimensional optimization problem as many different grasp candidates exist for a given object, with varying contact conditions and robot joint configurations. To simplify the problem, early works either assume prior knowledge of the object shape, e.g. via meshes or CAD models~\cite{bicchi2000robotic, bohg2014datadriven}, or they only consider top-down (i.e. 4-DoF) grasps~\cite{mahler2017dex, morrison2020learning}. In contrast, in this paper, we address the general case of full 6-DoF object grasping, where the exact object instance and category are unknown a priori, and only a single RGB-D image of the scene is observed.
% Describe main flaw of current state of the art: what can't we do yet?
Prior work has already shown impressive results in the general 6-DoF grasp-pose estimation problem~\cite {ten2017grasp, breyer2020volumetric, sundermeyer2021contact, jiang2021synergies}. Nevertheless, wider adoption beyond research labs is still limited. To close this gap, we identify two key features that the ideal object grasping pipeline should possess: 
\rebuttal{
\textit{holistic grasping}, i.e. the ability to predict grasps all around the given object, even in regions occluded from the agent's perspective, and \textit{object awareness}, i.e. the ability to identify different objects in the scene and distinguish between them as well as the background.
These properties would enable robots to grasp a specific target object, as opposed to just a random object from the clutter, as well as to reason about the best grasp poses independently of the current camera point of view, e.g. grasping a mug from the handle even if it is currently not visible.
}

To achieve \textit{holistic grasping}, prior works adopt a multi-camera setup around the scene~\cite{simeonovdu2021ndf}, while others require the agent to first execute a full scan of the scene prior to estimating grasps~\cite{breyer2020volumetric}. While these strategies are effective, they restrict the applicability of the method, since the former requires carefully instrumented environments, while the latter drastically reduces the speed of operation. Additionally, there are common scenarios where both methods are not applicable, for example, when looking inside a cabinet to grasp a mug.

\begin{figure}
    \centering
    \begin{subfigure}[b]{0.23\textwidth}
        \centering
        \includegraphics[trim=100 100 0 0,clip,width=\textwidth]{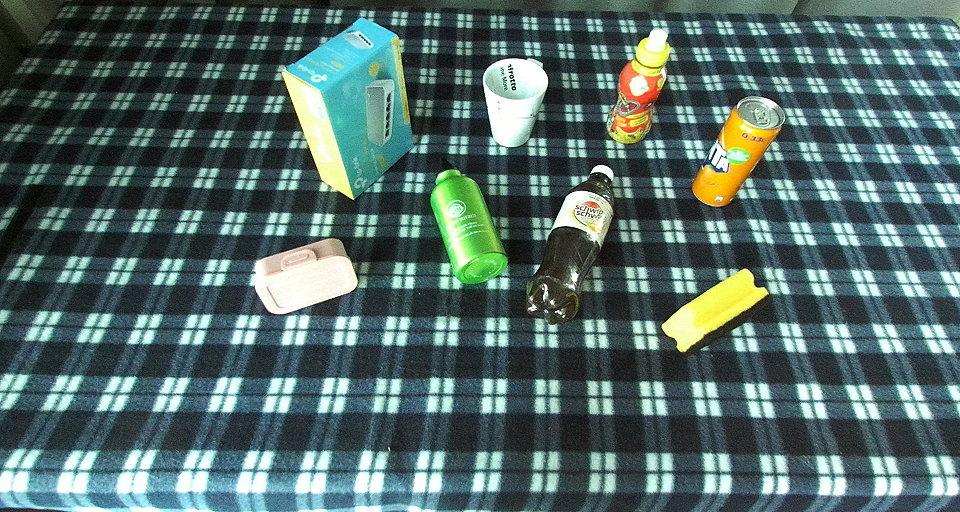}
        \vspace*{-6mm}
        \caption{\label{fig:main_rgb}}
    \end{subfigure}
    \vspace{1mm}
    \begin{subfigure}[b]{0.23\textwidth}  
        \centering 
        \includegraphics[trim=100 100 0 0,clip,width=\textwidth]{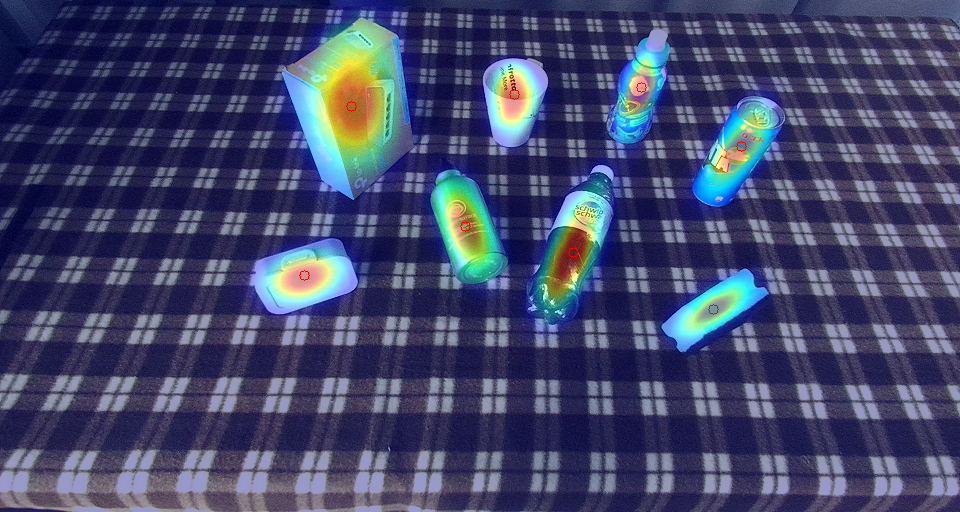}
        \vspace*{-6mm}
        \caption{\label{fig:main_heatmap}}
    \end{subfigure}
    \vspace{1mm}
    \begin{subfigure}[b]{0.23\textwidth}
        \centering
        \includegraphics[trim=100 100 0 0,clip,width=\textwidth]{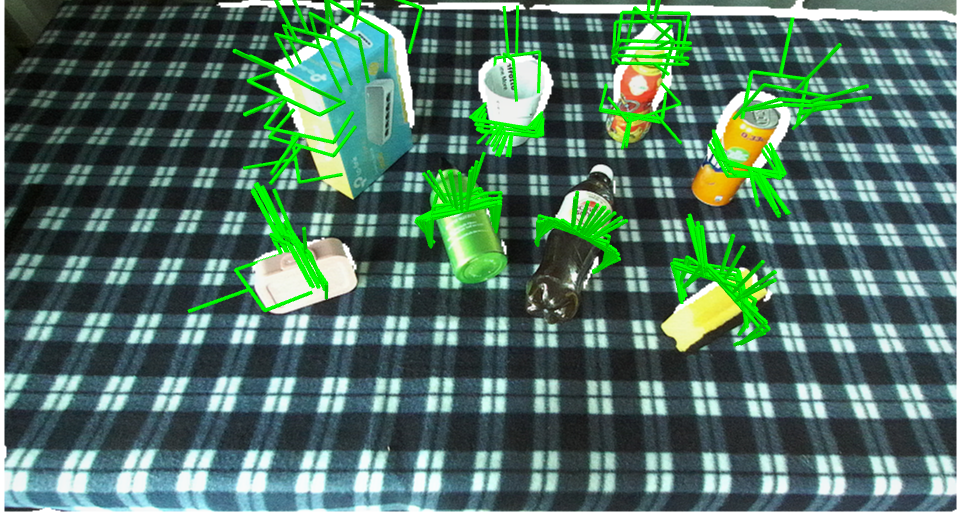}
        \vspace*{-6mm}
        \caption{\label{fig:main_overlay_multi}}
    \end{subfigure}
    \begin{subfigure}[b]{0.23\textwidth}   
        \centering 
        \includegraphics[trim=100 100 0 0,clip,width=\textwidth]{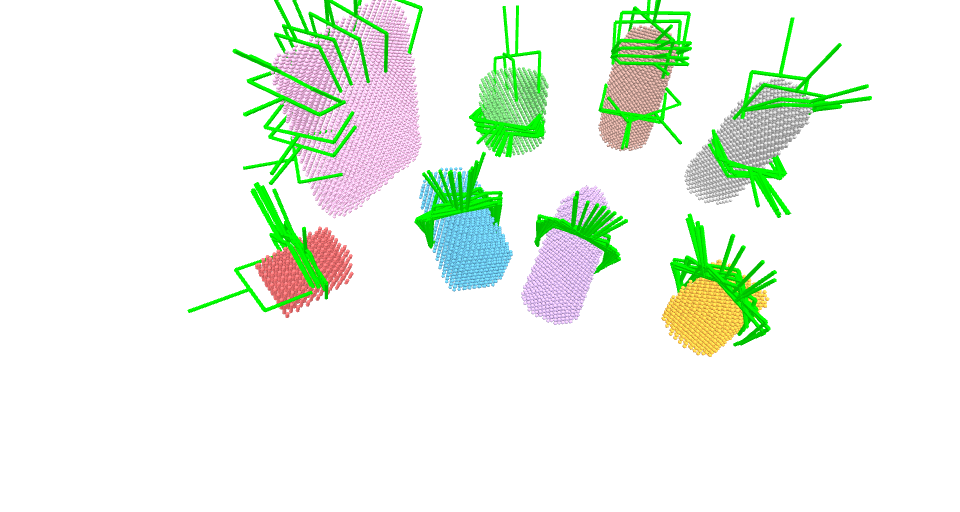}
        \vspace*{-6mm}
        \caption{\label{fig:main_reconstr_multi}}
    \end{subfigure}
    \begin{subfigure}[b]{0.23\textwidth}
        \centering
        \includegraphics[trim=100 100 0 0,clip,width=\textwidth]{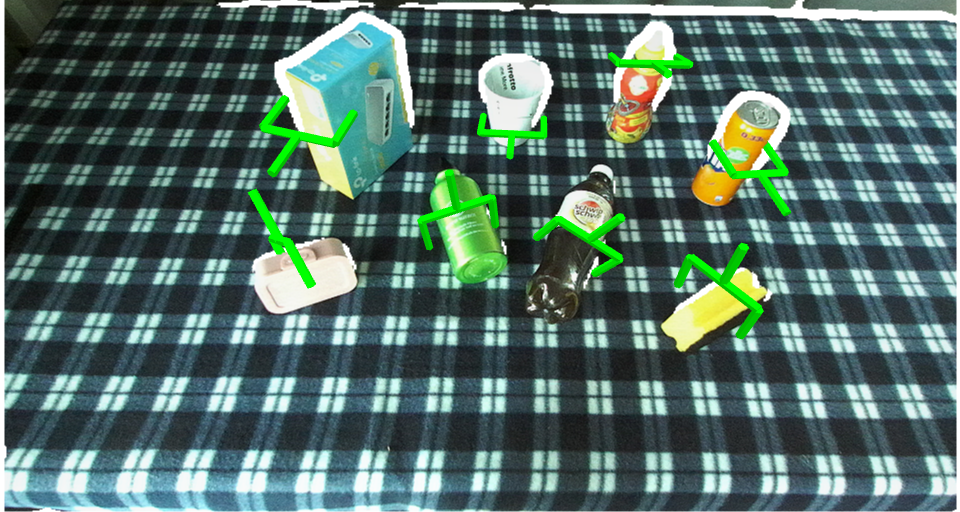}
        \vspace*{-6mm}
        \caption{\label{fig:main_overlay_single}}
    \end{subfigure}
    \begin{subfigure}[b]{0.23\textwidth}   
        \centering 
        \includegraphics[trim=100 100 0 0,clip,width=\textwidth]{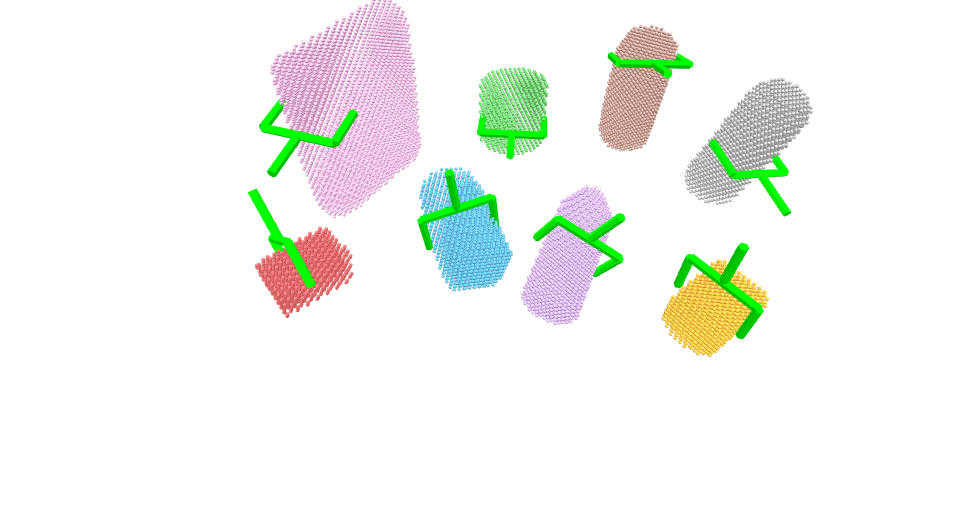}
        \vspace*{-6mm}
        \caption{\label{fig:main_reconstr_single}}
    \end{subfigure}
    \caption{{Overview of \ourName{}: 
        (\subref{fig:main_rgb}) Input image, 
        (\subref{fig:main_heatmap}) Object heatmap prediction, 
        (\subref{fig:main_overlay_multi}) Subset of all valid and final grasp proposals overlaid on the observed scene, 
        (\subref{fig:main_reconstr_multi}) Subset of all valid and final grasp proposals overlaid on the 3D scene reconstruction,
        (\subref{fig:main_overlay_single}) Selected grasps overlaid on the observed scene, 
        (\subref{fig:main_reconstr_single}) Selected grasps overlaid on the 3D scene reconstruction.
        }}
    \label{fig:main_fig}
    \vspace{-0.3cm}
\end{figure}

Most prior works are not object-aware, as they only consider the local geometry of the scene to predict grasps and discard additional information such as color images~\cite{breyer2020volumetric,jiang2021synergies}. 
Hence, these methods are limited to clutter removal tasks, where objects are grasped randomly from the scene. 
Recent work leveraged instance segmentation to achieve object awareness in their pipelines~\cite{sundermeyer2021contact}. However, two-step pipelines increase complexity during inference and are thus more error-prone. Further, a combined model has the advantage to exploit mutually beneficial information.
% Further, the two separate steps do not share mutually beneficial information.

% Describe our idea: the hero of the story comes to the rescue
In this work, we propose \ourName{}, a novel, single-shot object-aware holistic grasp prediction approach. It consists of an image encoder that predicts an objectness heatmap, a pose, and a latent code for each object, as well as an auto-decoder that reconstructs their respective 3D shape and predicts a set of grasp proposals.  We train \ourName{} only on synthetic images and demonstrate that it achieves zero-shot generalization to the real world.
% Provide a brief teaser of the results and summarize the contribution
We extensively test \ourName{} on both simulated as well as real-world scenes with two types of clutter: packed and pile. We evaluate on the joint task of simultaneous shape reconstruction and 6-DoF grasp estimation, and compare to the recent state-of-the-art method GIGA~\cite{jiang2021synergies}, demonstrating strong performance on this joint task.

This paper makes the following main contributions:
\begin{itemize} [noitemsep, topsep=-5px]
\item \ourName{}, a novel method for joint shape reconstruction and 6-DoF grasp estimation.
\item An automated 6-DoF robotic grasping evaluation environment based on the SAPIEN~\cite{Xiang_2020_SAPIEN} simulator.
\item Extensive experimental evaluations comparing \ourName{} with GIGA~\cite{jiang2021synergies}, a state-of-the-art baseline.
\item The code (including our dataset generation and evaluation script), models, and videos publicly available at \url{http://centergrasp.cs.uni-freiburg.de}.
\end{itemize}

% Nice Review Paper~\cite{newbury2022deep}, \\
% Acronym Dataset~\cite{acronym2020}, \\
% ContactGraspNet~\cite{sundermeyer2021contact}, \\
% GIGA~\cite{jiang2021synergies}

% Rebuttal additions: gou2021rgb, wang2021graspness, fang2023anygrasp, hoang2022context, hoang2023grasp, liu2023joint, chen2023efficient, zheng2022vgpn, qin2023rgb

\section{Related Work}
There has been a plurality of 6-DoF grasping works and object detection methods. In the following, we highlight recent advances in both topics.

\subsection{6-DoF Grasp Estimation}

The 6-DoF grasp-pose estimation problem has attracted significant interest over the years~\cite{newbury2022deep}. We identify two main 6-DoF grasp estimation schemes: point cloud-based and voxel-based.

{\parskip=3pt
\noindent\textit{Point cloud-based}: GPD~\cite{ten2017grasp} and PointNetGPD~\cite{liang2019pointnetgpd} are popular approaches where multiple grasp candidates are sampled using geometric heuristics from a point cloud, and a learned classifier is trained to evaluate their quality. 6-DoF GraspNet~\cite{mousavian20196} uses a variational autoencoder to predict grasps proposals from a single object point cloud.
\rebuttal{AVN~\cite{gou2021rgb} processes RGB and depth separately to be robust against depth noise. GSNet~\cite{wang2021graspness} and AnyGrasp~\cite{fang2023anygrasp} adopt a per-point graspness metric to highlight graspable areas and guide the prediction. VoteGrasp~\cite{hoang2022context, hoang2023grasp} uses contextual information to generate scene features that encode dependencies between objects. Liu~\textit{et~al.}~\cite{liu2023joint} propose a fusion network for joint foreground segmentation and grasp pose detection. Chen~\textit{et~al.}~\cite{chen2023efficient} present a grasp heatmap model to guide the prediction. 
% VPGN~\cite{zheng2022vgpn} proposes to aggregate geometric features from local regions via a voting mechanism. 
DGCAN~\cite{qin2023rgb} refines noisy depth features according to cross-modal relations with the RGB image.} All the aforementioned methods are purely geometrical and do not include the notion of an object. To tackle this problem, Contact-GraspNet~\cite{sundermeyer2021contact} proposes a two stages pipeline, where an unknown object segmentation model~\cite{xie2021unseen} is used to detect different object instances.
In contrast, our heatmap prediction inherently distinguishes between multiple objects in the scene and thus enables object-aware grasping.
}
% \noindent\textit{Point cloud-based}: Among early works, a popular approach is proposed by Pas and Platt~\cite{pas2015using}, where multiple grasp candidates are sampled using geometric heuristics from a point cloud, and a learned classifier is trained to evaluate their quality. It is later extended as GPD (Grasp Pose Detection)~\cite{ten2017grasp} and PointNetGPD~\cite{liang2019pointnetgpd}. More recently, 6-DoF GraspNet~\cite{mousavian20196} uses a variational autoencoder to predict grasps proposals from a single object point cloud. 
% All the aforementioned methods are purely geometrical and do not include the notion of an object. To tackle this problem, Contact-GraspNet~\cite{sundermeyer2021contact} proposes a two stages pipeline, where an unknown object segmentation model~\cite{xie2021unseen} is used to detect different object instances.
% In contrast, our heatmap prediction inherently distinguishes between multiple objects in the scene and thus enables object-aware grasping.
% }

{\parskip=3pt
\noindent\textit{Voxel-based}: Instead of point clouds, some methods opt to encode the observed scene in a Truncated Signed Distance Field (TSDF) volume~\cite{curless1996volumetric}.
Volumetric Grasping Network (VGN)~\cite{breyer2020volumetric} executes a scan motion with the robot's end-effector and integrates all camera depth observations in a TSDF voxel-grid, which is then used as input to predict grasps candidates. While effective, the scan motion drastically reduces the speed of operation of the robot.
To overcome this limitation, GIGA~\cite{jiang2021synergies} proposes to exploit the synergy between geometric reasoning and grasp learning. The key idea is that a representation capable of encoding the shape information of the scene is also relevant to grasp prediction and vice-versa. 
Nevertheless, voxel-based methods can only be applied to a workspace of fixed dimensions, whose coordinates and calibration with respect to the robot are known a priori. Contrary to point cloud-based methods, this requires additional setup efforts and it prevents their use in more dynamic settings, such as mobile manipulation~\cite{honerkamp2023n,schmalstieg2022learning}.}

\subsection{Object Detection and Representation}

In order to fulfill both \textit{object awareness} and \textit{holistic grasping}, a method must be able to both detect objects in the scene, as well as produce an appropriate representation from which grasps can be extracted.

{\parskip=3pt
\noindent\textit{Center-Based Detection}: The seminal work \emph{Objects as Points}~\cite{zhou_objects_2019} led to the development of many center-based object detection methods~\cite{duan_centernet_2019, heppert_category-independent_2022}. 
Recently, center-based detection methods have shown great results on category-level pose estimation and shape reconstruction~\cite{irshad_centersnap_2022, irshad2022shapo, lunayach2024fsd} with extension to articulated objects~\cite{heppert_carto_2023}. 
We build upon these single-shot frameworks, mitigating the issues of two-stage pipelines for grasping~\cite{manuelli_kpam_2019,sundermeyer2021contact}, and allowing us to predict a vast amount of grasp proposals while still retaining object-awareness.
}

{\parskip=3pt
\noindent\textit{Implicit Object Representation}: A useful object representation needs to be able to encode relevant geometric information and to generalize to unseen object instances. The recently proposed NUNOCS~\cite{wen2022catgrasp} represents each object as a 6D transformation and 3D scaling of the category "mean shape". This simple representation allows extracting dense correspondences and transferring relevant grasps across instances. While effective, this approach cannot handle large intra-category shape variations besides scaling. 
%To overcome this drawback, kPAM~\cite{manuelli_kpam_2019} proposes a novel semantic 3D keypoints formulation, which is able to handle a large variety of differently shaped object instances. However, this method requires manual keypoint labels, uses a separate instance segmentation pipeline, and needs to train a separate model for each object category. For both the aforementioned methods, the class of the object at inference time needs to be known. On the other hand, our method does not consider any explicit "category" information, avoiding this issue.
A different approach is to use the recently proposed OccupancyNet~\cite{mescheder2019occupancy} as an implicit representation. GIGA~\cite{jiang2021synergies} employs this model for scene reconstruction from partial observations. NDFs~\cite{simeonovdu2021ndf} adopt it to encode geometric and semantic descriptors in a 3D volume, which can then be queried to sample task-relevant keypoints~\cite{von2023treachery} for unseen instances. They show impressive results in terms of generalization to unseen object poses and shape instances, but they require a setup of four cameras, as well as a separate model for each category, and a set of manually collected demonstrations.
}

{\parskip=3pt
\noindent\textit{Grasp Distance Function}: DeepSDF~\cite{park_deepsdf_2019} is a learned continuous Signed Distance Function representation that implicitly encodes a surface as the zero-level set of the learned model. 
Neural Grasp Distance Fields (NGDF)~\cite{weng_neural_2022} extend the concept of the neural implicit distance function to grasping, where the learned model outputs scalar distance metrics for the current gripper pose to the closest grasp pose. 
Through differentiation, the joint positions of the robot can be iteratively updated, until the desired grasp pose is reached. 
While promising, this approach has only been evaluated with a single object in the scene and training a separate model for each category. 
Moreover, it only provides the gradient towards one single grasp pose (the closest one). 
On the other hand, \ourName{} is category-independent, it handles cluttered scenes, and it predicts an entire manifold of grasp candidates for each object in the scene.
Most similar to our method, SceneGrasp~\cite{agrawal2023real} also learns a combined latent space for shape and grasp prediction. While interesting, this approach has only been trained on 6 categories of objects, and it lacks evaluation of the grasp quality, both in simulation and in the real world.
}

\begin{figure*}
     \centering
      \includegraphics[width=\textwidth]{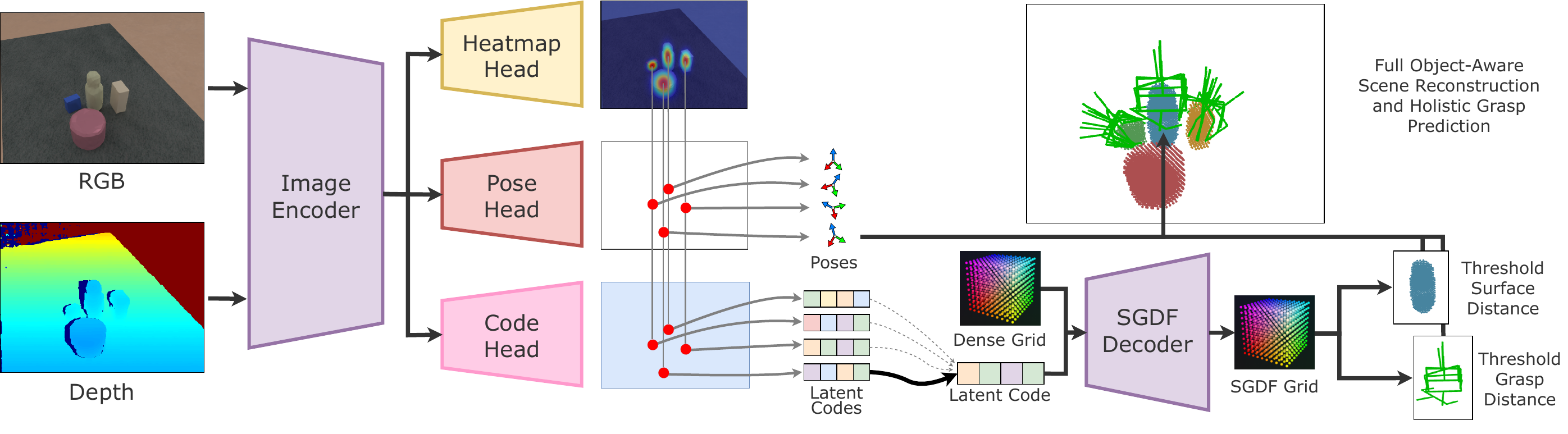}
     %\hspace{1.5cm}
     \caption{{Illustration of the \ourName{} architecture. First, an RGB-D image is fed into the image encoder which outputs an object heatmap, a pose map, and a latent code map. Next, the object locations in the image are determined by extracting the peaks from the predicted heatmap. At these locations, each object pose and latent code is extracted accordingly. In the second step, the SGDF decoder infers the shape and grasps for each detected object. Finally, the object pose is used to transform the shape and grasp predictions from the canonical frame to the camera frame.}}
    \label{fig:architecture}
    %\vspace{-0.25cm}
\end{figure*}

\section{Technical Approach}
\label{sec:technical_approach}

Given an RGB-D image as input, the goal is to simultaneously detect all the objects and predict their pose, shape, and valid grasps. 
We build upon the recently proposed CenterSnap~\cite{irshad_centersnap_2022} architecture, a single-shot 3D shape reconstruction and 6D pose estimation method, and adapt it to our grasp prediction setting. 
We adopt a point-based representation where we encode the complete 3D information of each object in the scene in the respective center point in the 2D image. 
At inference time, our image encoder (see \cref{section:rgb}) detects the center point of each object as well as the respective 6D pose and latent shape vector. 
This information is then used by the proposed shape and grasps distance function (SGDF) decoder discussed in \cref{section:sgdf} to reconstruct both 3D shapes and valid grasps.
Finally, as shown in \cref{fig:architecture}, we transform the 3D shape of the object and the grasps back to the camera frame using our predicted poses.

The \textit{object awareness} property is achieved thanks to the heatmap prediction, while the \textit{holistic grasping} property is obtained from the full object shape and grasps reconstruction capabilities of the SGDF decoder.
\rebuttal{The SGDF decoder is trained first to learn a latent embedding space. After training, the optimized latent codes for each object in the dataset are saved, so that they can be used as labels for training the latent code prediction of the image encoder.} 
In the following, we discuss both the encoder and decoder in detail as well as our data generation pipeline.

\subsection{Image Encoder}\label{section:rgb}

% {\parskip=3pt
{\noindent \textit{Model Architecture}:} Following the CenterSnap~\cite{irshad_centersnap_2022} architecture, we generate a low resolution feature representation of both the RGB and depth images using a ResNet50 %~\cite{he2016deep} 
topology. 
We then concatenate both representations and feed them into a ResNet18-FPN backbone~\cite{kirillov2019panoptic}. 
Finally, we use the resulting pyramid of features as input to three specialized heads: a heatmap head, a pose head, and a latent code head. All three network heads output a per-pixel prediction. The architecture of the three network heads follows the PanopticFPN~\cite{kirillov2019panoptic} segmentation head. It takes FPN features as input and fuses information from all levels into a single output. Each feature level is upsampled to a common dimension and then summed elementwise. Each upsampling stage consists of a $3\times3$ convolution, group norm, ReLU, and $2\times$ bilinear upsampling.
The heatmap head predicts the probability of objectness for each pixel as a scalar value $ o \in [0, 1]$. 
The pose head predicts the 6D pose of the object at the given pixel as a 12-dimensional vector, which is then reshaped into a $3\times4$ matrix and projected to a valid homogeneous transformation via a Procrustes operation~\cite{bregier2021deepregression}.
The code head predicts $\latentVariable \in \realNumber^{\latentVariablesDim}$, a $\latentVariablesDim$-dimensional vector representing the shape and possible grasps for the object at that pixel, denoted as latent code. 

{\parskip=3pt\noindent \textit{Training}:} 
Our heatmap loss $\lossGeneralFunction_\text{heat}$ is the mean squared error between the heatmap labels and the network output. 
Inspired by~\cite{sundermeyer2021contact}, we represent a 6D pose as a set of four 3D points $v \in \realNumber^{4\times3}$. This formulation allows us to define the pose loss as the simple Euclidean distance between points, unifying both translation and rotation information with a homogeneous metric.
Finally, the latent code loss $\lossGeneralFunction_\text{shape}$ is calculated as the L1 loss between prediction and ground truth code maps. 
Both the pose and latent code losses are masked by the ground truth object binary masks and weighted at each pixel by the ground truth value of the object heatmap. This ensures that no loss is computed on background pixels, and emphasizes higher weight towards the center of the objects. Our final, total loss is the weighted sum of these three losses, averaged over all pixels:
\begin{equation}
    \lossGeneralFunction_\text{encoder} = 
        \lossScaling_\text{heat} \lossGeneralFunction_\text{heat} +
        \lossScaling_\text{pose} \lossGeneralFunction_\text{pose} +
        \lossScaling_\text{shape} \lossGeneralFunction_\text{shape}.
\end{equation}

We report the scaling parameters in \cref{tab:hyper_parameters:encoder}.
We train the network for 100 epochs. To improve generalization, we normalize the RGB values to the dataset mean and standard deviation, and we add color jitter augmentation to the RGB images. We use the ADAM optimizer with default parameters and a learning rate of $1e-3$.
\begin{table}
    \centering
    \caption{Overview of hyperparameters.}
    \label{tab:hyper_parameters}
    \begin{subtable}[h]{0.22\textwidth}
        \centering
        \caption{Decoder}
        \label{tab:hyper_parameters:decoder}
        \begin{tabular}{l|p{0.5cm}c}
            \toprule
            Loss Scaling & Param. & Value  \\ 
            \midrule
            SDF & $\lossScaling_\text{SDF} $ & 10.0  \\
            Grasp & $\lossScaling_\text{Grasp} $ & 1.0 \\
            Code & $\lossScaling_\text{Code} $ & 0.001 \\
            \bottomrule
        \end{tabular}
    \end{subtable}
    \hspace{0.5cm}
    \begin{subtable}[h]{0.22\textwidth}
        \centering
        \caption{Encoder}
        \label{tab:hyper_parameters:encoder}
        \begin{tabular}{l|p{0.5cm}c}
            \toprule
            Loss Scaling & Param. & Value  \\ 
            \midrule
            Heatmap & $\lossScaling_\text{Heat} $ & 100  \\
            Shape & $\lossScaling_\text{Shape} $ & 1 \\
            Pose & $\lossScaling_\text{Pose} $ & 5 \\
            \bottomrule
        \end{tabular}
    \end{subtable}
\vspace{-0.25cm}
\end{table}

\subsection{Shape and Grasps Decoder}\label{section:sgdf}

\begin{figure}
    \centering
    % Old 
    % \includesvg[width=\textwidth]{images/latent_space_horizontal.svg}
    % \caption{{\small Visualization of the learned latent code space. On the left, we show a two-dimensional t-SNE projection of the resulting optimized latent codes for our training objects. To highlight that \ourName{} learned a continuous prior over the shapes and the grasp, we additionally show the reconstruction and grasps of the mean code for each category on the right.}
    % \vspace{-0.25cm}}
    % NH: All of this breaks ..
    % \includesvg[width=\textwidth]{images/latent_space_giga_objects.svg}
    % \includesvg[inkscapelatex=false]{images/latent_space_giga_objects.svg}
    \includegraphics[trim=0 40 0 40,clip,width=\columnwidth]{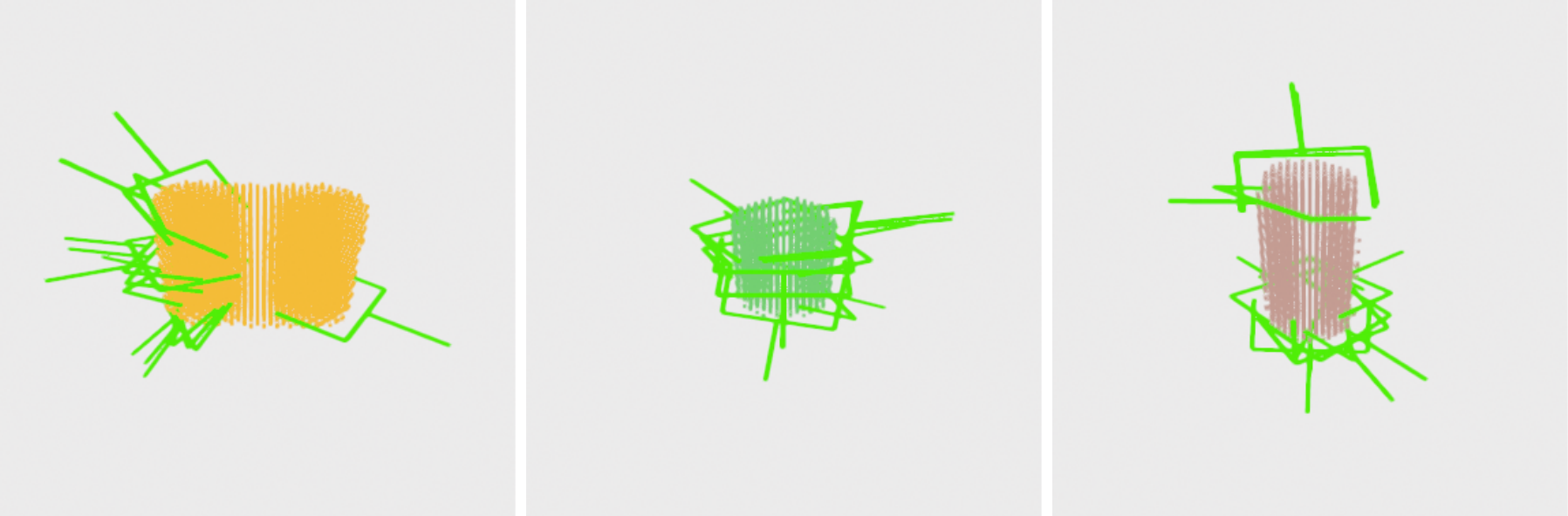}
    \caption{
        {Surface and Grasp Reconstruction. To highlight that \ourName{} learns a continuous prior over the shape and grasp manifold, we randomly sample latent codes from the learned embedding space and reconstruct the surface as well as ten valid grasps for each object.}
        %\vspace{-0.25cm}
    }
    \label{fig:latent_space}
\end{figure}

Our proposed shape and grasps distance function (SGDF) decoder maps a 3D coordinate and a latent code to both a shape distance and a grasp distance. The shape distance is a scalar value representing the signed distance between the sample coordinate and the closest point on the object's surface. 
The grasp distance is the full 6D pose of the closest valid grasp to the sample coordinate, expressed in a reference frame aligned with the object frame and centered at the sample coordinate itself. 

{\parskip=3pt\noindent \textit{Model Architecture}:} Our decoder network is a deep multi-layer perceptron based on DeepSDF \cite{park_deepsdf_2019}. 
The input to our decoder is a latent code $\latentVariable \in \realNumber^{\latentVariablesDim}$ and a coordinate in space $\spaceCoordinate \in \realNumber^{3}$. 
The decoder outputs a scalar value $\sdfValue \in \realNumber$ representing the distance to the closest surface as well as a grasp parameterized through a 9D vector raw grasp output $\graspOutput \in \realNumber^{9}$. 
The vector consists of three sub-vectors, an offset for the translation part of the grasp $\graspDelta \in \realNumber^{3}$ as well as two rotational components $\graspRotation_1, \graspRotation_2 \in \realNumber^{3}$. 
We build a true, absolute rotation matrix defined in the object frame $\graspRotationFull \in \realNumber^{3 \times 3}$ by applying the Gram-Schmidt procedure on the two rotational components. 
Thus, the final, full grasp pose is defined by constructing a transformation matrix using both components $\grasp = \transformation \left( \spaceCoordinate + \graspDelta, \graspRotationFull \right) \in SE(3)$ where the matrix is constructed through
\begin{equation}
    \transformation \left( \boldsymbol{t}, \boldsymbol{R} \right) = 
    \begin{bmatrix}
         \boldsymbol{R} &  \boldsymbol{t} \\
        \boldsymbol{0} & 1
    \end{bmatrix},
\end{equation}
where $\boldsymbol{t} \in \realNumber^{3}$ and $\boldsymbol{R} \in \realNumber^{3 \times 3}$.

{\parskip=3pt\noindent \textit{Training}:} We train the decoder using multiple losses: first, we impose a loss on clamped SDF values 
\begin{equation}
    \lossGeneralFunction_\text{SDF} = 
        \frac
        {            
            \sum_{
                \sdfValue_\text{gt}, \sdfValue_\text{pred} \in \allSdfValues
            }
            \left| 
                \text{clamp} \left( \sdfValue_\text{gt}, \clampThreshold \right)
                -
                \text{clamp} \left( \sdfValue_\text{pred}, \clampThreshold \right) 
            \right|
        }{
            | \allSdfValues |
        },
\end{equation}
where $\allSdfValues$ is the set of all ground truth $\sdfValue_\text{gt}$ and predicted $\sdfValue_\text{pred}$ SDF values. The clamping function 
\begin{equation}
    \text{clamp} \left( x, \clampThreshold \right) = 
        \min ( \max ( x , -\clampThreshold ), \clampThreshold )
\end{equation}
which clamps a value $x$ to be between the specified range. In the experiments, we used $\clampThreshold=0.1$

Second, we use a 5-point grasp pose loss similar to \cite{sundermeyer2021contact}. We use five  3D points on the gripper defined in the gripper frame, make them homogeneous, and stack them.
We denote the resulting matrix as $\gripperPoints \in \realNumber^{4 \times 5}$. To accommodate for symmetry we flip the points along the x-y-plane, resulting in $\gripperPointsFlipped \in \realNumber^{4 \times 5}$. 
We then calculate the grasping loss at each point as

{\footnotesize
\begin{equation}
    \lossGeneralFunction_\text{Grasp} = \frac
    {
        \sum_{\grasp_\text{gt}, \grasp_\text{pred} \in \allGrasps} \min 
        \left(
            \| \grasp_\text{gt} \gripperPoints - \grasp_\text{pred} \gripperPoints \|,
            \| \grasp_\text{gt} \gripperPointsFlipped - \grasp_\text{pred} \gripperPoints \|
        \right)
    }{
        n_{g}
    },
\end{equation}}
where we multiply the ground truth grasp $\grasp_\text{gt}$ with the original and flipped gripper point matrices $\gripperPoints$ and $\gripperPointsFlipped$, and use the minimal distance between either of them. The result is averaged over the number of grasps in the batch $n_g$.
Third, we regularize the latent codes through a loss $\lossGeneralFunction_\text{Code}$ computed as
\begin{equation}
    \lossGeneralFunction_\text{Code} = \bar{\allLatentVariables} 
    \cdot \min (1, \epoch / 5),
    % \begin{cases}
    %     \epoch / 5 & \text{if } \epoch < 5 \\
    %     1 & \text{else}
    % \end{cases}
\end{equation}
where $\bar{\allLatentVariables}$ is the average norm of all embeddings and $\epoch$ is the current training epoch.

We calculate our final loss as
\begin{equation}
    \lossGeneralFunction_\text{decoder} = 
        \lossScaling_\text{SDF} \lossGeneralFunction_\text{SDF} +
        \lossScaling_\text{Grasp} \lossGeneralFunction_\text{Grasp} +
        \lossScaling_\text{Code} \lossGeneralFunction_\text{Code}.
\end{equation}
We report the scaling parameters in \cref{tab:hyper_parameters:decoder}. Throughout all of our experiments, we set the latent code size $\latentVariablesDim$ to 32.

We train the network for 100 epochs. We apply dropout with probability 0.2 and weight normalization to regularize training. We use the ADAM optimizer with default parameters and a learning rate of $1e-3$.
\cref{fig:latent_space} shows the reconstructed output of our SGDF function for different sampled latent codes.

\subsection{Inference}
\label{sec:model_inference}

\begin{figure}[t]
     \centering
     \begin{subfigure}[b]{0.155\textwidth}
         \centering
         \includegraphics[trim=200 100 200 100,clip,width=\textwidth]{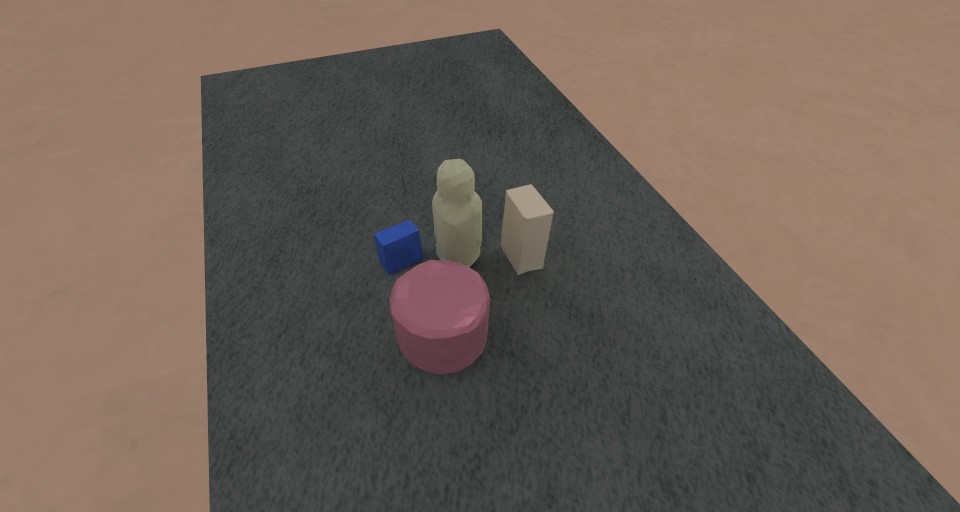}
         % \caption{RGB Image}
         % \label{subfig:synth_data_rgb}
     \end{subfigure}
     \begin{subfigure}[b]{0.155\textwidth}
         \centering
         \includegraphics[trim=200 100 200 100,clip,width=\textwidth]{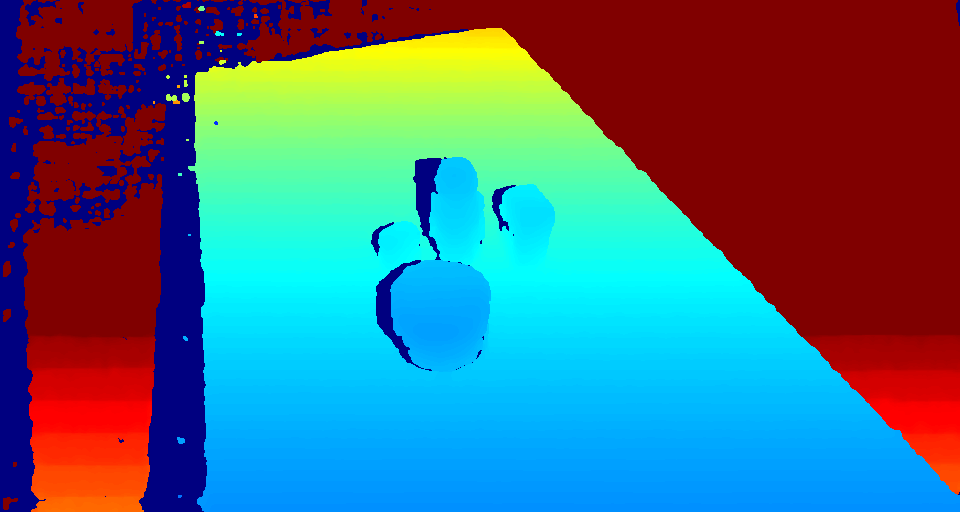}
         % \caption{Depth Image}
         % \label{subfig:synth_data_depth}
     \end{subfigure}
     % \begin{subfigure}[b]{0.10\textwidth}
     %     \centering
     %     \includegraphics[width=\textwidth]{images/poses_sim.png}
     %     \caption{GT Poses}
     %     \label{subfig:synth_data_poses}
     % \end{subfigure}
     \begin{subfigure}[b]{0.155\textwidth}
         \centering
         \includegraphics[trim=200 100 200 100,clip,width=\textwidth]{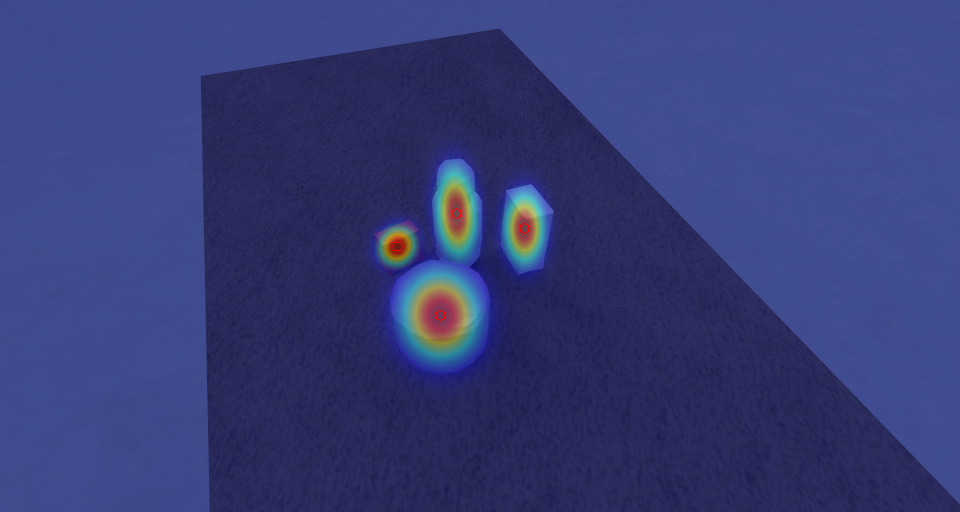}
         % \caption{GT Heatmap}
         % \label{subfig:synth_data_heatmap}
     \end{subfigure}
        \caption{Generated Synthetic Data. To generate our training data, we render a random scene, consisting of a floor, table, and between 1 to 5 objects to yield an RGB image (left), a depth image with simulated sensor noise (center), 
        % (\subref{subfig:synth_data_poses}) canonical ground truth poses transformed into the in-camera frame, 
        and an object heatmap (right).
        }
        %\vspace{-0.25cm}}
        \label{fig:synth-data}
\end{figure}

Given a single RGB-D observation of the scene, our image encoder (see \cref{section:rgb}) predicts a corresponding object heatmap, pose map, and latent code map. For each heatmap peak, both the 6D pose and latent code are extracted. For each object, we input to the SGDF decoder a $64\times64\times64$ dense grid of sampled 3D coordinates together with the predicted latent code. We then extract the object surface points and a manifold of valid grasps as the respective iso-surfaces \(SGDF(\cdot) = 0\), i.e.\ by discarding all points and grasps above a small threshold \( \epsilon \).
The remaining points and grasps are then transformed from their canonical frame to the camera frame using the predicted 6D object pose. 

To achieve higher precision, this pose is further refined via point-to-plane ICP to match the predicted shape with the observed point cloud. The list of predicted grasps is then filtered to discard all grasps that are in collision with the observed scene. Of the remaining grasp poses, we select the one that minimizes the torque around the grasp point, estimated using the gravitational force acting on the centroid of the predicted shape. These poses are then transformed from the camera frame into the robot base frame and the grasp is executed via an inverse kinematics controller.

\subsection{Data Generation}
\label{subsec:data_gen}

\ourName{} is trained solely on synthetic data and achieves zero-shot generalization to the real world. For a fair comparison, we train on the same 99,945 training scenes from GIGA~\cite{jiang2021synergies}, which consist of a list of object meshes and their respective poses. They are constructed from a set of 417 meshes as either packed or piled clutter.

{\parskip=3pt\noindent\textit{Shape and Grasp Distance Labels}:} 
To generate grasp candidates for each object mesh, we first sample 1,000 surface points from the object. For each point, we align the gripper fingers to the point normal. 
Finally, we sample 24 different wrist rotations uniformly, \rebuttal{i.e. we consecutively apply a 15 degrees rotation around the z-axis of the gripper,} for a total of 24,000 grasp candidates per object. 
From these candidates, we compute the ground truth grasp labels by checking for collisions and evaluating their antipodality. On average, we find around 8,500 valid grasps per object.
Similarly to NGDF~\cite{weng_neural_2022}, we find that a large discrete grasp set is a good approximation of the continuous manifold of valid grasps. 
For each object, we use \verb|mesh_to_sdf|\footnote{\href{https://github.com/marian42/mesh_to_sdf}{https://github.com/marian42/mesh\_to\_sdf}} to sample 100,000 points and their respective SDF values. 
For each point, we then find the closest grasp from the set of ground truth grasps of the given object and compute the respective grasp distance label.

{\parskip=3pt\noindent\textit{Image Labels}:} To generate synthetic image observations (RGB and depth) as well as pose and instance segmentation labels from the precomputed GIGA scenes, we use the raytracing-based renderer from SAPIEN~\cite{Xiang_2020_SAPIEN} and its realistic depth feature~\cite{zhang2023close}. 
All the textures, materials, lights, and table shapes are randomized, see \cref{fig:synth-data} for an example observation. For training, we render each scene from two random camera poses, resulting in roughly 200,000 RGB-D images and labels.
To generate the object heatmaps, whose peaks represent each object in the image, we fit a Gaussian to the ground truth masks.
For the pose map and code map labels, we use the instance masks to label each pixel of an object with its ground truth pose vector (12-dimensional) and its ground truth code vector ($\latentVariablesDim$-dimensional) respectively.

\section{Experimental Results}

\begin{table*}[t]
\centering
\caption{Evaluation of the shape and grasp-pose prediction in four different simulated environments. Results are averaged over three random seeds. \textit{\mbox{CD = L2-Chamfer Distance ($\downarrow$) in mm}, \mbox{IoU = 3D Intersection over Union ($\uparrow$)}, \mbox{SR = Success Rate ($\uparrow$)}, \mbox{DR = Declutter Rate ($\uparrow$)}.} }
\label{tab:sim-results}
\begin{tabularx}{2\columnwidth}{l c c c c c c c c c c c c c c c c c c c}
\cmidrule{1-20}
& \multicolumn{4}{c}{GIGA Objects Packed} && \multicolumn{4}{c}{GIGA Objects Pile} && \multicolumn{4}{c}{YCB Objects Packed} && \multicolumn{4}{c}{YCB Objects Pile} \\ 
\cmidrule{1-20}
& CD & IoU  & SR & DR && CD & IoU  & SR & DR && CD & IoU  & SR & DR && CD & IoU  & SR & DR \\
\cmidrule{2-5} \cmidrule{7-10} \cmidrule{12-15} \cmidrule{17-20}
GIGA~\cite{jiang2021synergies} 
& 35.5 & 0.18 & 0.52 & 0.59 && 71.5 & 0.10 & 0.41 & 0.18 && 63.3 & 0.13 & 0.49 & 0.41 && 83.2 & 0.11 & 0.44 & 0.27 \\
CG w/o ICP 
& 55.5 & 0.10 & 0.23 & 0.16 && 55.3 & 0.20 & 0.17 & 0.12 && 63.2 & 0.13 & 0.14 & 0.14 && 87.9 & 0.14 & 0.18 & 0.14  \\
CenterGrasp
& \textbf{23.2} & \textbf{0.34} & \textbf{0.84} & \textbf{0.87} && \textbf{16.1} & \textbf{0.66} & \textbf{0.84} & \textbf{0.71} && \textbf{18.1} & \textbf{0.61} & \textbf{0.83} & \textbf{0.81} && \textbf{42.0} & \textbf{0.47} & \textbf{0.67} & \textbf{0.70}  \\ 
\cmidrule{1-20}
\end{tabularx}
\end{table*}

\newcommand{\subimage}[1]{
\begin{minipage}{0.154\textwidth}
  \centering
  \includegraphics[width=\linewidth]{#1}
\end{minipage}%
}

\newcommand{\ylabel}[1]{
\begin{minipage}{0.02\textwidth}
\centering
\begin{tikzpicture}
  \node[rotate=90] {\footnotesize #1};
 \end{tikzpicture}
\end{minipage}%
}

\newcommand{\xlabel}[1]{
\begin{minipage}{0.15\textwidth}
\centering
\begin{tikzpicture}
  \node[] {\footnotesize #1};
 \end{tikzpicture}
\end{minipage}%
}

\begin{figure*}[t]
\ylabel{RGB}
\subimage{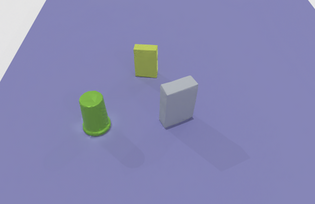}
\subimage{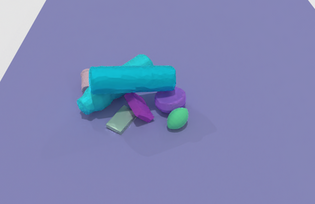}
\subimage{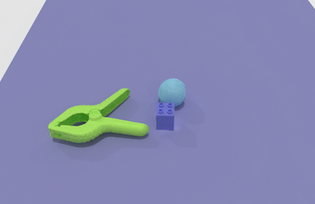}
\subimage{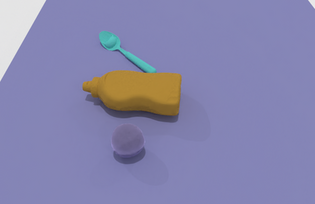}
\subimage{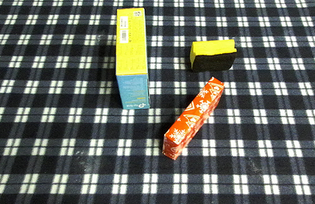}
\subimage{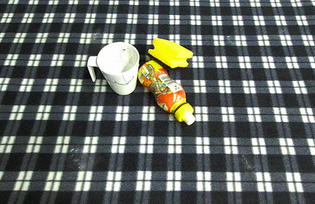}

\smallskip
\ylabel{GIGA}
\subimage{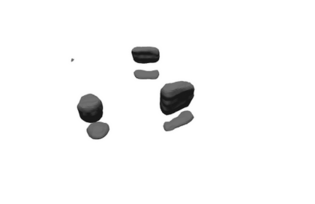}
\subimage{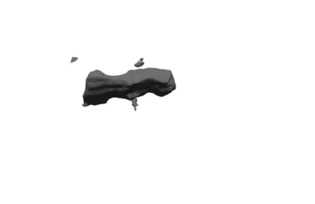}
\subimage{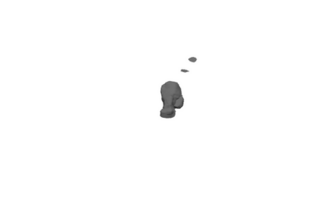}
\subimage{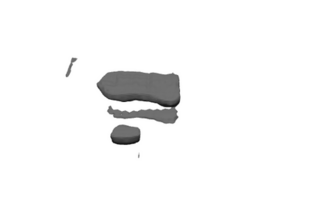}
\subimage{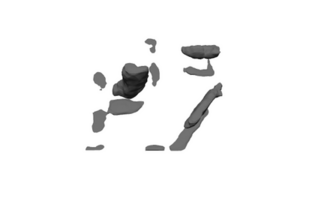}
\subimage{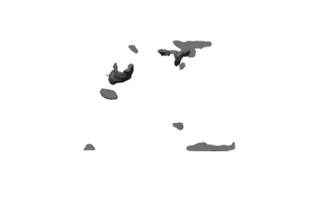}

\smallskip
\ylabel{CenterGrasp}
\subimage{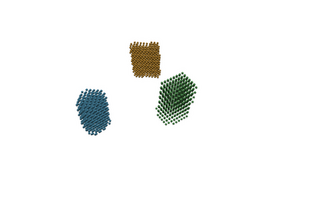}
\subimage{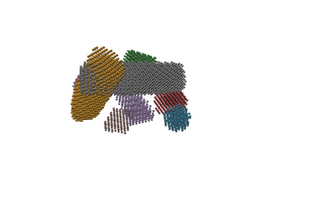}
\subimage{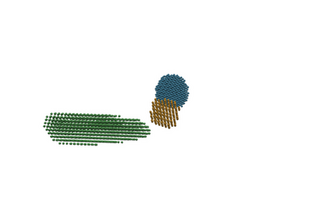}
\subimage{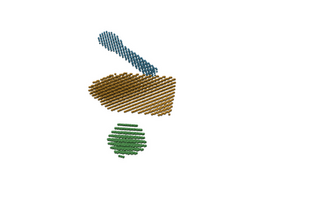}
\subimage{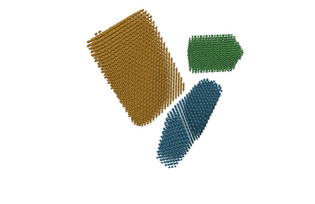}
\subimage{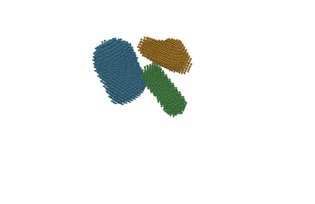}

\ylabel{ }
\xlabel{Giga Objects Packed}
\xlabel{Giga Objects Pile}
\xlabel{YCB Objects Packed}
\xlabel{YCB Objects Pile}
\xlabel{Real Objects Packed}
\xlabel{Real Objects Pile}
\caption{A comparison between the mesh reconstruction of GIGA~\cite{jiang2021synergies} and the point cloud reconstruction of \ourName{}. GIGA yields adequate results in the in-distribution evaluation (i.e. using GIGA objects), but its reconstruction quality drastically decreases in the out-of-distribution settings. On the other hand, \ourName{} demonstrates good reconstruction quality in all environments, including real-world evaluations.
}
% \vspace{0.5cm}
\label{fig:detection_reconstruction}
\end{figure*}

In this section, we first evaluate the proposed \ourName{} on the joint task of simultaneous shape reconstruction and 6-DoF grasp estimation. We present a comparison with the state-of-the-art baseline GIGA~\cite{jiang2021synergies}, as well as an ablation study of our \ourName{} approach without the ICP pose refinement step described in \cref{sec:model_inference}. 
It is important to note that the ICP refinement is only possible thanks to the \textit{object awareness} property of our approach: predicting separate shapes for each object allows us to refine their poses individually. This is not possible with GIGA, since its prediction is at the full scene level.
\rebuttal{Moreover, we evaluate the output grasp quality on the GraspNet-1Billion~\cite{fang2020graspnet} benchmark that evaluates grasp proposals via an analytic model}.
Finally, we demonstrate zero-shot transfer to a real robot setup.

\subsection{Simulation Experiments}

We build our evaluation tasks on SAPIEN~\cite{Xiang_2020_SAPIEN}, a realistic and physics-rich simulator. Our scenes consist of a Franka Emika Panda arm, a table, and a number of objects placed on top of it. While GIGA~\cite{jiang2021synergies} evaluated the grasping performance with a floating gripper that can be placed freely in space, we evaluate with a complete 7-DoF robot arm, which more closely mirrors a real-world scenario. To ensure a fair comparison with GIGA~\cite{jiang2021synergies}, we first use the same object meshes and randomized scenes from their test set. Additionally, we evaluate the out-of-distribution capability of all methods by testing on the YCB object set~\cite{calli2015benchmarking}. Following prior works~\cite{breyer2020volumetric, jiang2021synergies}, we generate two different types of cluttered scenes: packed clutter, where objects are randomly placed upright next to each other, and pile clutter, where objects are randomly dropped on the table and lie on top of each other. From the combination of two object sets and two clutter types, our full evaluation consists of four different environments. Each baseline is evaluated on 200 random scenes from each environment. The number of objects in each scene is sampled from a Poisson distribution with an expected value equal to 4.

{\parskip=3pt\noindent\textit{Shape Reconstruction}:} To evaluate the shape reconstruction quality, we use the ground truth meshes and their poses in the scene to extract a ground truth point cloud. 
We then compare it to both the point cloud predicted by \ourName{} as well as to a point cloud sampled from GIGA's predicted meshes. 
\cref{fig:detection_reconstruction} shows a qualitative comparison of the reconstruction from both methods. GIGA yields adequate results in the in-distribution evaluation (i.e. using GIGA objects), but its reconstruction quality drastically decreases in the out-of-distribution settings. On the other hand, \ourName{} demonstrates good reconstruction quality in all environments, including real-world evaluation.

In~\cref{tab:sim-results}, we report the quantitative results of our evaluation, which consists of two shape reconstruction metrics calculated between predictions and ground truth: the L2 bi-directional Chamfer Distance (CD), and the 3D Intersection over Union (IoU).
Averaging across environments, \ourName{} achieves an improvement of \SI{38.5}{\milli\meter} in CD and $0.38$ in the IoU score.
This numerical evaluation is consistent with the qualitative results from \cref{fig:detection_reconstruction}, highlighting the better generalization capabilities of \ourName{} compared to the state of the art. These results demonstrate the advantages of learning to predict geometries at the object level rather than at the full scene level.
Furthermore, our ablation of \ourName{} without ICP demonstrates the importance of pose refinement. The results show that \ourName{} without ICP achieves similar results to GIGA when evaluated with YCB objects, and worse results when evaluated with GIGA objects.

{\parskip=3pt\noindent\textit{Grasp Pose Prediction}:} We perform comprehensive evaluations of the grasping capabilities of \ourName{} on our simulation setup. For each scene, the robot executes the predicted grasp, lifts the object, and moves it to a drop area. The attempt is considered successful if the objects are successfully moved to the drop area. The episode is terminated when all objects in the scene have been cleared or after three consecutive failed attempts.
In \cref{tab:sim-results}, we report two object grasping metrics: Success Rate (SR) and Declutter Rate (DR). Success rate is defined as the number of successful grasps divided by the total grasps attempted, whereas declutter rate is defined as the number of successful grasps divided by the total number of objects in the evaluation scenes. 
The latter metric is important to highlight cases where objects are still present in the scene, but no grasp is predicted by the model.

In this experiment, \ourName{} achieves 33 percentage points higher SR and 41 percentage points higher DR compared to GIGA on average, demonstrating its ability to clear drastically more objects from the scene while requiring fewer attempts.
Furthermore, our ablation study shows a significant decrease in the success rate when using \ourName{} without ICP, which achieves significantly worse results than both \ourName{} and GIGA. Compared to the full pipeline, \ourName{} without ICP achieves over 60 percentage points less SR and DR, highlighting the importance of precise object poses for the success of the entire pipeline.

\subsection{GraspNet-1Billion benchmark}

\rebuttal{
While the target task of our work is simultaneous shape reconstruction and grasp estimation, we additionally evaluate the grasp quality of our method on the GraspNet-1Billion benchmark~\cite{fang2020graspnet}, to allow a direct comparison with dedicated grasp-pose estimation approaches. 
This benchmark evaluates the top 50 grasp proposals using an analytic model of grasp quality. 
Following~\cite{fang2020graspnet}, we report the Average Precision (AP) for two different friction coefficients ($\text{AP}_{0.4}$ and $\text{AP}_{0.8}$), as well as the total AP averaged across the full range of coefficients. 
We train and evaluate \ourName{} on the Kinect camera data. The results are shown in Tab.~\ref{tab:graspnet1b}, where we compare against the baselines originally reported in the benchmark.
For the seen objects case, \ourName{} performs better than the two lowest scoring baseline~\cite{morrison2018closing, chu2018real} and worse than the two best scoring methods~\cite{liang2019pointnetgpd, fang2020graspnet}. For the unseen and novel object cases, a similar trend is observable for the $\text{AP}_{0.4}$ metric. We can attribute this underperformance to multiple factors. First, \ourName{} is designed to detect objects in the scene, extract their poses and predict their full shape and grasp manifold, while the compared methods only focus on the grasp estimation task considered in this benchmark. Second, the benchmark reports the average metric over the best 50 grasps: since \ourName{} does not directly predict a grasp score, ranking the poses in the grasp manifold is not straightforward. Third, we observe that the heatmap objectness prediction struggles in heavily cluttered scenes in the benchmark, leading to failure cases downstream. Fourth, the small amount of objects in the GraspNet-1B dataset (only 87), paired with a high variability in their shape, prevents the SGDF decoder to generalize well to unseen objects.
}

\begin{table*}[t]
\centering
\caption{\rebuttal{Evaluation of 6-Dof grasp pose prediction on the GraspNet-1Billion benchmark~\cite{fang2020graspnet}. We report AP for two friction coefficients (0.4, 0.8) as well as the total AP averaged across the full range of coefficients. \textit{AP = Average Precision ($\uparrow$), G = Grasp, R = Reconstruction, D = Detection.}}}
\label{tab:graspnet1b}
% \begin{tabularx}{1.8\columnwidth}{>{\color{blue}}l c c c c | c >{\color{blue}}c >{\color{blue}}c >{\color{blue}}c >{\color{blue}}c >{\color{blue}}c >{\color{blue}}c >{\color{blue}}c >{\color{blue}}c >{\color{blue}}c >{\color{blue}}c >{\color{blue}}c >{\color{blue}}c}
\begin{tabularx}{1.8\columnwidth}{l c c c c | c c c c c c c c c c c c c}
\cmidrule{1-17}
Method &&&&&& \multicolumn{3}{c}{Seen} && \multicolumn{3}{c}{Unseen} && \multicolumn{3}{c}{Novel} \\ 
\cmidrule{1-17}
& G & R & D &&& $\text{AP}$ & $\text{AP}_{0.8}$ & $\text{AP}_{0.4}$ && $\text{AP}$ & $\text{AP}_{0.8}$ & $\text{AP}_{0.4}$ && $\text{AP}$ & $\text{AP}_{0.8}$ & $\text{AP}_{0.4}$ \\
\cmidrule{2-4} \cmidrule{7-9} \cmidrule{11-13} \cmidrule{15-17}
GG-CNN~\cite{morrison2018closing} 
& \checkmark & - & - &&& 16.89 & 22.47 & 11.23 && 15.05 & 19.76 & 6.19 && 7.38 & 8.78 & 1.32\\
Chu et al.~\cite{chu2018real} 
& \checkmark & - & - &&& 17.59 & 24.67 & 12.74 && 17.36 & 21.64 & 8.86 && 8.04 & 9.34 & 1.76 \\
GPD~\cite{ten2017grasp} 
& \checkmark & - & - &&& 24.38 & 30.16 & 13.46 && 23.18 & 28.64 & 11.32 && 9.58 & 10.14 & 3.16 \\
PointnetGPD~\cite{liang2019pointnetgpd} 
& \checkmark & - & - &&& 27.59 & 34.21 & 17.83 && 24.38 & 30.84 & 12.83 && 10.66 & 11.24 & 3.21\\
GraspNet-1B~\cite{fang2020graspnet} 
& \checkmark & - & - &&& \textbf{29.88} & \textbf{36.19} & \textbf{19.31} && \textbf{27.84} & \textbf{33.19} & \textbf{16.62} && \textbf{11.51} & \textbf{12.92} & \textbf{3.56} \\ 
\cmidrule{1-17}
CenterGrasp
& \checkmark & \checkmark & \checkmark &&& 22.35 & 27.16 & 15.35 && 13.78 & 17.01 & 8.22 && 5.72 & 6.93 & 2.64 \\
\cmidrule{1-17}
\end{tabularx}
\vspace{-0.3cm}
\end{table*}

\subsection{Real Robot Experiments}
\label{sec:real-robot}

 \begin{figure}
    \centering
     \begin{subfigure}[c]{0.23\textwidth}
         \centering
         \includegraphics[width=\textwidth]{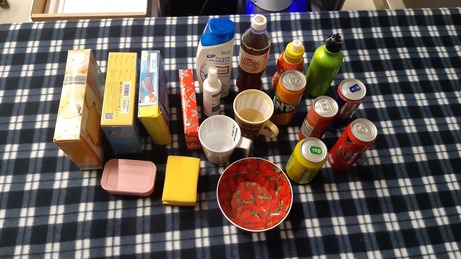}
         % \caption{\label{fig:robot}}
     \end{subfigure}
     \begin{subfigure}[c]{0.18\textwidth}
         \centering
         \includegraphics[width=\textwidth]{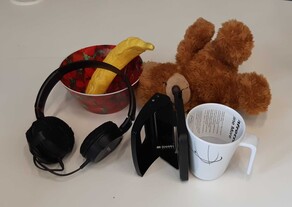}
         % \caption{\label{fig:objects}}
     \end{subfigure}
   \caption{The simple (left) and non-convex (right) household objects used in our real-world experiments.}
   \label{fig:experiment-platform}
 \end{figure}

 \begin{table}[t]
\setlength{\tabcolsep}{3.5pt} % make space between columns smaller
\centering
\caption{{Evaluation of 6-Dof grasp pose prediction in the real world. \textit{SR = Success Rate ($\uparrow$), DR = Declutter Rate ($\uparrow$), G = Grasp, R = Reconstruction, D = Detection.}}}
\label{tab:real-results}
\begin{tabularx}{\columnwidth}{l c c c c | c c c c c c c c c}
\cmidrule{1-14}
Method &&&&&& \multicolumn{2}{c}{Packed} && \multicolumn{2}{c}{Pile} && \multicolumn{2}{c}{NonConvex} \\ 
\cmidrule{1-14}
& G & R & D &&& SR & DR  && SR & DR && SR & DR\\
\cmidrule{2-4} \cmidrule{7-8} \cmidrule{10-11} \cmidrule{13-14}
\rebuttal{GPD}~\cite{ten2017grasp}
& \checkmark & - & - &&& \rebuttal{0.48} & \rebuttal{0.79} && \rebuttal{0.41} & \rebuttal{0.67} && \rebuttal{0.38} & \rebuttal{0.59} \\
GIGA~\cite{jiang2021synergies} 
& \checkmark & \checkmark & - &&& 0.36 & 0.64 && 0.20 & 0.37 && \rebuttal{0.26} & \rebuttal{0.44} \\
CenterGrasp
& \checkmark & \checkmark & \checkmark &&& \textbf{0.57} & \textbf{0.85} && \textbf{0.53} & \textbf{0.83} && \rebuttal{\textbf{0.49}} & \rebuttal{\textbf{0.70}} \\ 
\cmidrule{1-14}
\end{tabularx}
\end{table}

We carry out evaluations of the grasp prediction quality on a real robot setup consisting of a 7-DoF Franka Emika Panda arm equipped with a parallel jaw-gripper and a wrist-mounted ZED2 depth camera.
We collect a set of 20 simple \rebuttal{and a set of 6 non-convex household objects, shown in~\cref{fig:experiment-platform}. We evaluate on three settings: packed and pile (similar to the simulation experiments) and non-convex. We build 30 different scenes with three objects each, for a total of 90 possible grasps for all three scenarios}.
To minimize variance, we replicate the scenes across all the methods as accurately as possible. 
For each scene, we capture an RGB-D image and query the model to predict a grasp. We then execute the predicted grasp and remove the object from the scene. 
If a grasp attempt fails two times for the same object, that object is removed manually from the scene.
For each baseline and type of scene, we report the success rate (SR) and declutter rate (DR) in~\cref{tab:real-results}.
Similar to the simulation results, we observe that \ourName{} demonstrates the best performance across all the evaluations, achieving on average 27 percentage points higher SR and 20 percentage points higher DR compared to GIGA.

\section{Discussion and Limitations}

 \begin{figure}[t]
    \centering
     \begin{subfigure}[t]{0.15\textwidth}
         \centering
         \includegraphics[trim=400 200 400 200,clip,width=\textwidth]{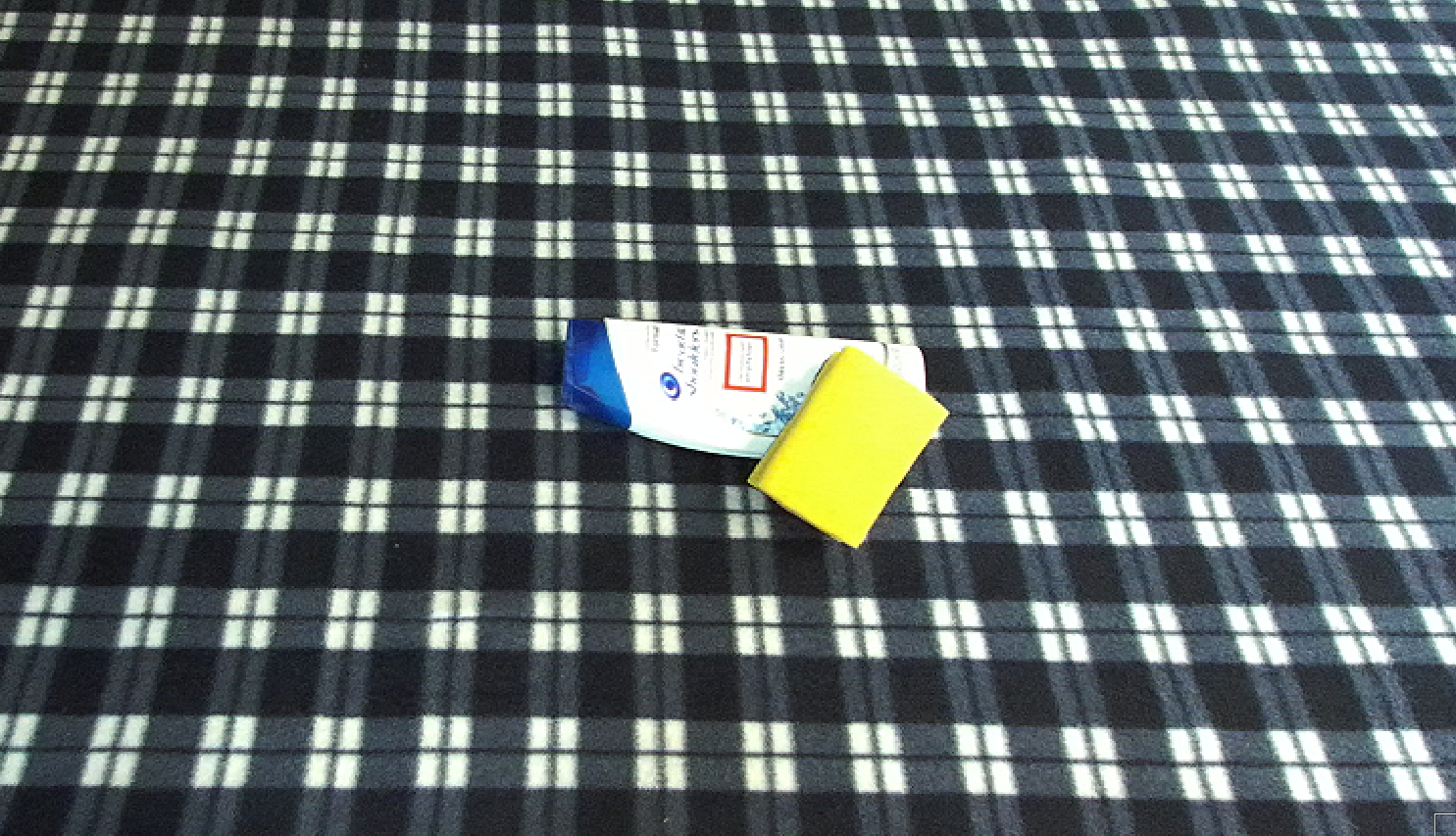}
         % \caption{RGB image.}
         % \caption{}
         % \label{fig:partial_object_rgb}
     \end{subfigure}
     ~
     \begin{subfigure}[t]{0.15\textwidth}
         \centering
         \includegraphics[trim=600 200 600 200,clip,width=\textwidth]{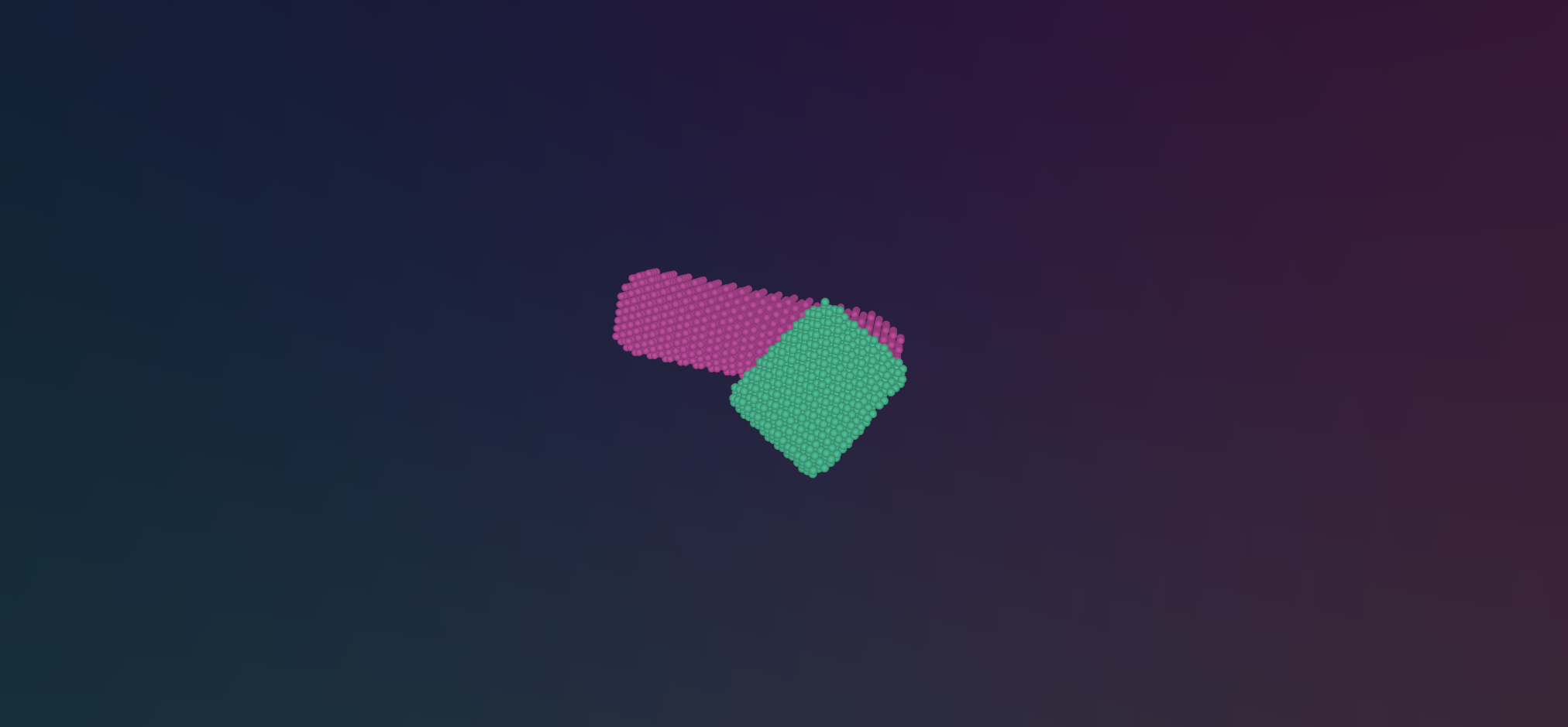}
         % \caption{Reconstruction viewed from camera view. }
         % \caption{}
         % \label{fig:partial_object_camera_view}
     \end{subfigure}
     ~
     \begin{subfigure}[t]{0.15\textwidth}
         \centering
         \includegraphics[trim=600 200 600 200,clip,width=\textwidth]{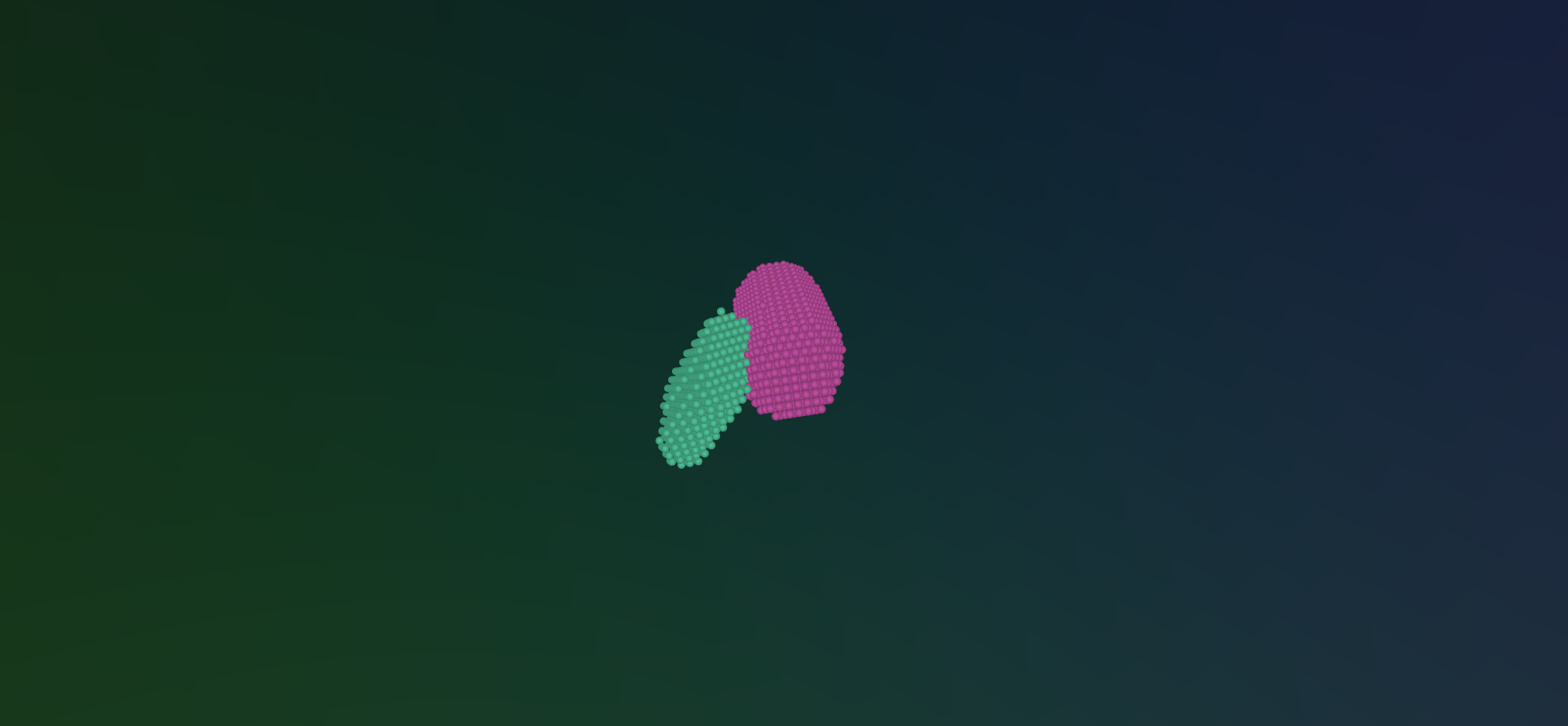}
         % \caption{Reconstruction rotated $90\deg$. \label{fig:partial_object_rotated_view}}
         % \caption{}
         % \label{fig:partial_object_rotated_view}
     \end{subfigure}
   \caption{Object-aware and holistic scene understanding. Given the input image (left), \ourName{} is able to predict a full per-object reconstruction even in the presence of partial occlusions (center-right).}
   \label{fig:scene_understanding}
 \end{figure}

\begin{figure} %
% \begin{minipage}[t]{\columnwidth}
    \centering
    % \begin{subfigure}[t]{0.45\columnwidth}
    \begin{minipage}{0.48\linewidth}
    % \begin{figure}[t]%{0.45\linewidth}
        \centering
        \includegraphics[width=\columnwidth]{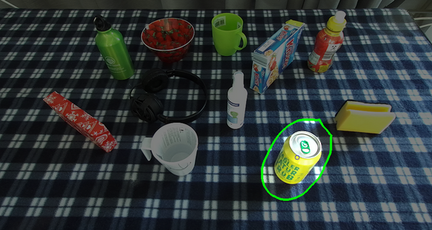}
        \captionof{figure}{The \textit{object awareness} property allows human interaction. User can select the desired object to be grasped.}
        \label{fig:object_select}
    % \end{subfigure}
    \end{minipage}
    % \end{figure}
    ~
    % \begin{subfigure}[t]{0.45\columnwidth}
    \begin{minipage}{0.48\linewidth}
    % \begin{figure}[t]%{0.45\linewidth}
        \centering
        \includegraphics[width=\columnwidth]{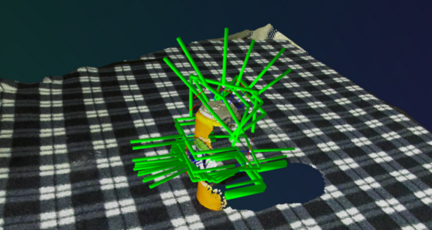}
        \captionof{figure}{The \textit{holistic} grasping property allows \ourName{} to predict grasps even behind the visible region of the camera (viewing from left to right).}
        \label{fig:holistic_grasping}
    % \end{subfigure}
    \end{minipage}
    % \end{figure}
%     \caption{\nick{}}
%     \label{fig:centergrasp_properties}
% \nick{I made this two seperate figures side, by side}
% \vspace{0.5cm}
% \end{minipage}
\end{figure}

The effective scene understanding of \ourName{} not only enhances grasping capabilities but also unlocks various additional opportunities. For instance, the concept of \textit{object awareness} enables human interaction, allowing users to select desired objects by utilizing object masks extracted from the heatmap, as shown in \cref{fig:object_select}. 
Additionally, the ability of \ourName{} to perform \textit{holistic grasping} ensures accurate detection, reconstruction, and successful grasping of objects, even when they are partially (self-) occluded (\cref{fig:scene_understanding}), or behind the visible region of the camera (\cref{fig:holistic_grasping}).
\rebuttal{For example, with the object in \cref{fig:holistic_grasping}, we predict a total of 1267 grasps, of which 61\% are on the visible region and 39\% on the invisible region of the object.}
Although our approach demonstrates significant performance improvements compared to the state-of-the-art baseline, it does have some limitations. 
First, precise pose predictions are critical to the success of the pipeline, which in practice implies the adoption of an ICP refinement step.
Second, two separate models need to be trained, the image encoder, which operates at the scene level, and the SGDF decoder, which operates at the object level.
Both aspects represent an exciting opportunity for future research.

\section{Conclusion}

In this work, we presented \ourName{}, a simultaneous shape reconstruction and grasp-pose prediction architecture. We designed our framework around two key concepts: \textit{object awareness} and \textit{holistic grasping}, with the aim to combine 3D scene understanding with object grasping. We compared the proposed approach with a recent state-of-the-art method and showed that it achieves exceptional performance both in simulation and real-world robotic experiments. By combining accurate perception and successful grasping, our framework paves the way for dexterous robotic systems capable of effectively interacting with objects in complex real-world environments. We made the code and trained models publicly available to facilitate future research.

%===============================================================================

%===============================================================================
{
\footnotesize
\bibliographystyle{IEEEtran}
\bibliography{IEEEabrv,centergrasp.bib}
}

% \appendix

% % Reset Counters
% \setcounter{section}{0}
% % \setcounter{equation}{0}
% % \setcounter{figure}{0}
% \setcounter{table}{0}
% % \setcounter{page}{1}
% \makeatletter
% \renewcommand{\thesection}{S.\arabic{section}}
% \renewcommand{\thesubsection}{S.\arabic{section}.\arabic{subsection}}
% \renewcommand{\thetable}{S.\arabic{table}}
% \renewcommand{\thefigure}{S.\arabic{figure}}
% \renewcommand{\theequation}{S.\arabic{equation}}

% \normalsize
% \input{supplementary/01_Model_Details.tex}

\end{document}